\documentclass[11pt]{article}

\usepackage[margin=1in]{geometry}
\usepackage{amsmath,amssymb,amsthm}
\usepackage{booktabs}
\usepackage{tikz}
\usetikzlibrary{positioning,arrows.meta,calc}
\usepackage[font=small,labelfont=bf]{caption}
\usepackage[hidelinks]{hyperref}
\usepackage{url}
\usepackage{authblk}

\title{Why Conclusions Diverge from the Same Observations:\\
Formalizing World-Model Non-Identifiability via an Inference Profile \texorpdfstring{$\theta$}{theta}}

\author{Toru Takahashi}
\affil{Human Informatics and Systems Laboratory, Doshisha University, Kyoto, Japan\\
Linked Open Data Initiative, NPO, Tokyo, Japan\\
Keio Research Institute at SFC, Fujisawa, Japan\\
Stroly Inc., Kyoto, Japan\\
\texttt{toru@mis.doshisha.ac.jp}}

\date{}

\begin{document}
\maketitle

\begin{abstract}
When people share the same documents and observations yet reach different conclusions, the disagreement often shifts into a judgment that the other party is cognitively defective, irrational, or acting in bad faith. This paper argues that such divergence is better described as a form of non-identifiability inherent in inference and learning, rather than as a defect of the other party. We organize the phenomenon into two levels: (i) $\theta$-level non-identifiability, where conclusions diverge under the same world model $W$ because inference settings differ; and (ii) $W$-level non-identifiability, where repeated use of an inference setting $\theta$ biases data exposure and update rules, causing the learned world model $W$ itself to diverge. We introduce an inference profile $\theta = (R, E, S, D)$, consisting of Reference, Exploration, Stabilization, and Horizon, and show how outputs can split even for the same observation $o$ and the same $W$. We further explain why disagreements tend to project onto a small number of bases---abstract versus concrete, externalizability, and order versus freedom---as a consequence of general constraints on learning systems: computational, observational, and coordination constraints. Finally, we relate the framework to deep representation learning, including representation hierarchy, latent-state estimation, and regularization--exploration trade-offs, and illustrate the framework through a case study on AI regulation debates.
\end{abstract}

\section{Introduction}

People often examine the same documents, statistics, logs, or incidents and nevertheless reach different conclusions. This phenomenon appears repeatedly in policy making, organizational decision making, research and development, product safety, and public debates on emerging technologies. What is especially problematic is not only that conclusions diverge, but that such divergence is often reinterpreted as a defect of the other party: lack of ability, lack of sincerity, ideological corruption, or moral deficiency. Once disagreement is framed in this way, the discussion rarely moves toward productive activities such as additional observation design, model comparison, or refinement of inference procedures.

The aim of this paper is to separate conclusion divergence from personality evaluation and to formulate it as a problem of non-identifiability in world-model estimation. A world model is understood here as a statistical and computational device that forms internal representations from observation sequences in order to support prediction and decision making. \emph{Non-identifiability} refers to a situation in which, under finite data, partial observability, and representational constraints, multiple models or inference policies remain compatible with the same observations, so that the conclusion is not uniquely determined by the observations alone~\cite{rothenberg1971,lewbel2019}.

This paper distinguishes two levels of non-identifiability.

\begin{itemize}
\item \textbf{$\theta$-level non-identifiability}: even under the same world model $W$, conclusions may diverge if inference settings differ, including which grounds are adopted, how widely alternatives are explored, when updating stops, and which temporal horizon is emphasized.
\item \textbf{$W$-level non-identifiability}: repeated use of an inference setting $\theta$ biases exposure to data and update rules over time, causing the learned world model $W$ itself to diverge. Once this happens, the same observation is processed through different causal and representational structures.
\end{itemize}

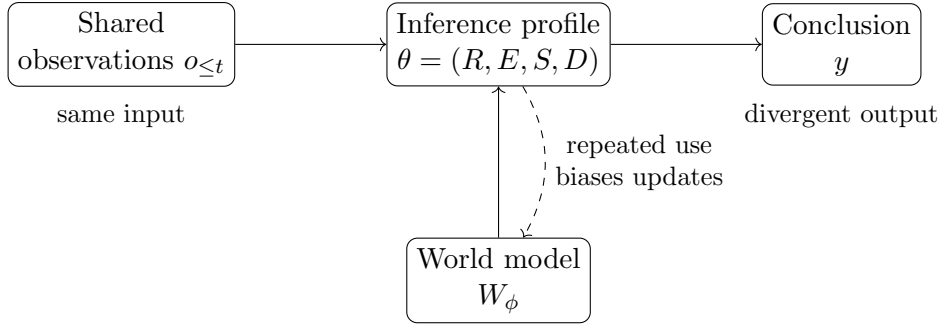
\begin{figure}[h]
\centering
\begin{tikzpicture}[node distance=2cm,every node/.style={align=center}]
\node[draw,rounded corners] (obs) {Shared\\observations $o_{\leq t}$};
\node[draw,rounded corners,right=of obs] (theta) {Inference profile\\$\theta = (R,E,S,D)$};
\node[draw,rounded corners,right=of theta] (y) {Conclusion\\$y$};
\node[draw,rounded corners,below=of theta] (w) {World model\\$W_\phi$};
\draw[->] (obs) -- (theta);
\draw[->] (theta) -- (y);
\draw[->] (w) -- (theta);
\draw[->,dashed] (theta) to[bend left=30] node[right,font=\small] {repeated use\\biases updates} (w);
\node[below=0.1cm of obs,font=\small] {same input};
\node[below=0.1cm of y,font=\small] {divergent output};
\end{tikzpicture}
\caption{Two levels of non-identifiability. At the $\theta$ level, different inference profiles can produce different conclusions even under the same world model. At the $W$ level, repeated inference operations bias data exposure and updates, causing the world model itself to diverge.}
\label{fig:two-level}
\end{figure}

This two-level structure makes it possible to treat both short-term misalignment and long-term epistemic fragmentation within a single framework. Figure~\ref{fig:two-level} gives a schematic overview.

The contribution of this paper is threefold. First, it introduces an inference profile $\theta = (R, E, S, D)$ as a compact representation of inference operation. Second, it explains why this four-dimensional operational freedom tends to be projected onto three recurrent bases of disagreement: abstract/concrete, externalizability, and order/freedom. Third, it connects these bases to structural features of deep representation learning, thereby grounding the framework in computational terms rather than treating disagreement merely as a matter of values or communication style.

The central claim is not that all disagreements are equally valid, nor that inference differences should be normatively celebrated. Rather, the claim is that disagreement often has an identifiable computational structure. By locating divergence in inference operation and world-model learning, we can shift the problem from moralized evaluation to designable coordination.

\section{Background: SIA and MIA}

\subsection{The Single Intelligence Assumption}

When people disagree despite sharing observations, many discussions implicitly rely on what this paper calls the \emph{Single Intelligence Assumption} (SIA). SIA is not a single formal doctrine but a composite of three assumptions.

\begin{enumerate}
\item \textbf{Centralization}: the essence of intelligence is assumed to lie in abstract symbol manipulation or logical reasoning, while other modes of inference are treated as secondary.
\item \textbf{Normativization}: the centralized form of intelligence is treated as the uniquely correct norm, so deviations from it are interpreted as insufficient effort, insufficient education, or ethical failure.
\item \textbf{Commutability}: if the same input is provided, properly rational agents are expected to arrive at the same conclusion.
\end{enumerate}

SIA can be useful in education, evaluation, and institutional operation because it provides a common standard. However, the stronger the commutability assumption becomes, the more likely divergence is to be attributed to defects of the other party rather than to differences in inference operation, observation design, or model structure. This attribution blocks inquiry into observation design, inference settings, and model mismatch.

\subsection{The Multiple Inference Assumption}

As an alternative, this paper adopts the \emph{Multiple Inference Assumption} (MIA). MIA treats diversity in inference operation and learning not as an exception to be eliminated, but as a consequence of non-identifiability in world-model estimation. The key point is not to normatively celebrate diversity, but to describe divergence as a set of adjustable design degrees of freedom rather than as bad faith or cognitive deficiency.

MIA therefore functions as a methodological assumption: when the same observation leads to different conclusions, the first question should be where the inference process differs, and whether the difference belongs to an operational setting $\theta$ or to a learned world model $W$.

\subsection{Externalizability and Internalization}

This paper characterizes grounds and states in terms of the cost of sharing and verification across agents.

\begin{itemize}
\item \textbf{Externalizability}\footnote{An alternative phrasing is \emph{objectifiability}: the degree to which a state can be described as an object that exists beyond any particular subject. We use \emph{externalizability} throughout this paper to emphasize the operational aspect of inter-agent sharing and verification, rather than the metaphysical question of whether the state has subject-independent existence.}: the degree to which a ground or state can be described in a form that is shareable, reproducible, and auditable across agents. Measurements, logs, legal texts, and benchmark scores have high externalizability.
\item \textbf{Internalization}: the degree to which a state is inferred as an internal state of an agent, such as confidence, anxiety, expectation, tacit knowledge, or a sense of danger. Such states cannot be uniquely reconstructed from observation $o$ and typically require high description cost to be communicated.
\end{itemize}

This distinction is not an epistemological opposition between objective and subjective. It is an operational distinction concerning communication cost and verifiability under coordination constraints.

\section{Problem Setting: Two Levels of Non-Identifiability}

\subsection{Minimal Representation of World Models and Inference}

An observation $o$ is not a single value. It may consist of documents, statistics, logs, cases, reports, and temporal sequences of such evidence. An inference agent forms a latent state $z_t$ from an observation sequence $o_{\leq t}$ and outputs a conclusion $y_t$~\cite{bengio2013}. We write a world model as
\begin{equation}
W_\phi : o_{\leq t} \mapsto (z_t, p_\phi(y_t \mid o_{\leq t})),
\label{eq:world-model}
\end{equation}
where $\phi$ denotes the parameters of the model.

The phenomenon that conclusions diverge from the same observations can be written as a situation in which, for agents $A$ and $B$, the conclusions generated from the same observation sequence are not identical:
\begin{equation}
y_A \neq y_B \quad \text{even though} \quad o^A_{\leq t} = o^B_{\leq t}.
\label{eq:divergence}
\end{equation}

\subsection{\texorpdfstring{$\theta$}{theta}-Level Non-Identifiability}

$\theta$-level non-identifiability occurs when conclusions diverge because inference operation differs even under the same world model. Inference is not merely a function of the observation and model parameters; it also includes operational settings such as reference selection, exploration width, update inhibition, and temporal horizon:
\begin{equation}
y = \mathrm{Infer}(W_\phi, o_{\leq t}; \theta).
\label{eq:infer}
\end{equation}
Even if $\phi$ is held fixed, the conclusion need not be uniquely determined if agents differ in which grounds they prioritize, which alternatives they maintain, how easily they update, or over what future interval they evaluate consequences.

\subsection{\texorpdfstring{$W$}{W}-Level Non-Identifiability}

Inference operations are repeated. Repetition affects which observations are selected, which additional data are collected, which update rules are activated, and how future outcomes are discounted. Thus the data-generating process experienced by an agent may depend on $\theta$:
\begin{equation}
o_t \sim P(\cdot \mid \theta_t), \qquad \phi_{t+1} = U(\phi_t, o_t, \theta_t),
\label{eq:update}
\end{equation}
where $U$ is an update function.

Consequently, even when two agents appear to be looking at the same news, documents, or incidents, the long-term data-generating processes they have experienced may differ. Their learned world models may therefore diverge, producing different causal attributions, different expectations about intervention, and different judgments about what counts as a relevant signal.

\section{Inference Profile \texorpdfstring{$\theta = (R, E, S, D)$}{theta = (R, E, S, D)}}

This section defines four operational degrees of freedom in inference. The purpose is to locate conclusion divergence not in vague differences of atmosphere or personality, but in identifiable points of the inference process.

\subsection{Reference \texorpdfstring{$R$}{R}: Weighting of Grounds}

When many grounds are available, inference depends on which ones are adopted. In coordination environments, grounds that can be explained and audited by others tend to be prioritized.

Assume that an observation $o$ can be decomposed into a set of partial grounds $\{e_i\}_{i=1}^{N}$, such as statistical values, log fragments, rules, or cases. Let $\psi(e_i)$ be the representation of each ground. The reference representation is a weighted composition:
\begin{equation}
r = \sum_{i=1}^{N} w_i \psi(e_i), \qquad y = F_\theta(W_\phi, r).
\label{eq:reference}
\end{equation}
The first source of divergence is therefore the difference in the weights $w_i$.

For each ground $e_i$, introduce an externalizability score $x_i \in [0, 1]$. Let $c(e_i)$ be the minimum description length required to translate $e_i$ into a form that is auditable across agents. We define
\begin{equation}
x_i = \exp(-\alpha c(e_i)),
\label{eq:externalizability}
\end{equation}
where $\alpha > 0$. Grounds such as measurements, logs, and legal texts have high $x_i$, whereas intuitions, tacit knowledge, and difficult-to-verbalize concerns have low $x_i$.

A simple form of reference policy is given by
\begin{equation}
w_i \propto \exp(\beta_R x_i + u_i),
\label{eq:reference-policy}
\end{equation}
where $u_i$ includes content-based compatibility terms and $\beta_R$ controls the degree to which externalizable grounds are prioritized. A large $\beta_R$ shifts inference toward logs, statistics, and other auditable grounds. A small $\beta_R$ allows high-description-cost grounds (such as tacit concerns or unarticulated risk perception) to enter the inference process.

\subsection{Exploration \texorpdfstring{$E$}{E}: Retention of Alternative Hypotheses}

Inference may converge early to a single conclusion or retain multiple possibilities. Let conclusions be represented as hypotheses $h \in \mathcal{H}$ and let $p(h \mid o)$ be the posterior distribution over hypotheses. Exploration controls how many alternatives remain active.

The entropy of the hypothesis distribution characterizes this degree of freedom:
\begin{equation}
H(h \mid o) = -\sum_{h} p(h \mid o) \log p(h \mid o).
\label{eq:entropy}
\end{equation}
A strong exploration setting keeps $H(h \mid o)$ high; a weak one lowers entropy by concentrating on a small number of hypotheses~\cite{cover2006}. In generative models, temperature and top-$p$ sampling play analogous roles. In human inference, a related contrast appears between fast, intuitive processing and slower, deliberative processing~\cite{kahneman2011}.

\subsection{Stabilization \texorpdfstring{$S$}{S}: Update Inhibition and Stopping Conditions}

When new information arrives, an inference system may flexibly revise its judgment or preserve an established rule. Let $\eta$ be an online update parameter for an inference or decision rule, and let $\Delta\eta$ be the proposed update. A simple stopping condition is
\begin{equation}
\text{update if } |\Delta\eta| > \tau.
\label{eq:stopping}
\end{equation}
A large threshold $\tau$ makes the policy harder to change, producing order and reproducibility. A small threshold $\tau$ makes the policy more responsive to new information. Stabilization can also be expressed through regularization strength $\lambda$ in the update rule, where stronger regularization produces greater stability.

\subsection{Horizon \texorpdfstring{$D$}{D}: Temporal Center of Evaluation}

Decision making depends on whether short-term outcomes or long-term consequences are emphasized. Let $\gamma \in [0, 1]$ be a discount factor. A larger $\gamma$ places greater weight on long-term consequences, whereas a smaller $\gamma$ emphasizes immediate response and local adaptation~\cite{sutton2018}. Horizon $D$ therefore represents the temporal center of evaluation.

\section{Projection onto Three Bases}

\subsection{Three General Constraints}

Why do the degrees of freedom in $\theta$ often appear as a small number of recurrent axes? This section explains the reduction in terms of three constraints common to learning systems.

\begin{enumerate}
\item \textbf{Computational constraints} $C_{\mathrm{comp}}$: finite representation capacity, inference time, and learning resources.
\item \textbf{Observational constraints} $C_{\mathrm{obs}}$: partial observability, noise, and uncertainty in the data-generating process.
\item \textbf{Coordination constraints} $C_{\mathrm{coop}}$: accountability, reproducibility, auditability, and speed of agreement formation.
\end{enumerate}

\subsection{Three Trade-Offs Implied by the Constraints}

\paragraph{Compression trade-off: abstract versus concrete.}
Under $C_{\mathrm{comp}}$, representation must be compressed. Rate--distortion theory shows that retaining input detail increases representation cost, whereas stronger compression increases reconstruction distortion~\cite{cover2006}. The degree to which a system preserves concrete details or moves toward abstract generalization is therefore not uniquely determined.

\paragraph{Externalization trade-off: externalizable versus internalized grounds.}
Under $C_{\mathrm{obs}}$, agents must estimate hidden states that cannot be uniquely recovered from observation. Under $C_{\mathrm{coop}}$, however, grounds are expected to be shareable and auditable. Externalizing internal states requires communication cost. Differences in tolerance for this cost generate an axis between externalizable and internalized grounds.

\paragraph{Stability trade-off: order versus freedom.}
Learning systems face a trade-off between update inhibition and adaptation to novelty. This is closely related to the plasticity--stability dilemma~\cite{grossberg1987,mccloskey1989}. Stabilization and exploration form an axis between reproducible order and retained diversity. Information-theoretically, this axis can be understood as the amount of uncertainty the system is allowed to maintain. Order corresponds to lower entropy and reproducibility; freedom corresponds to higher entropy and retained alternatives.

\subsection{Relation between \texorpdfstring{$\theta$}{theta} and the Three Bases}

The four components of $\theta$ tend to be projected onto three bases under the above constraints.

\begin{itemize}
\item \textbf{Reference $R$} projects onto the externalization trade-off. Under coordination constraints, the dominant degree of freedom is whether highly externalizable grounds are prioritized, controlled by $\beta_R$.
\item \textbf{Exploration $E$ and Stabilization $S$} jointly project onto the order/freedom trade-off. Exploration entropy and stopping thresholds are coupled by computational and coordination constraints.
\item \textbf{Horizon $D$} projects onto the abstract/concrete trade-off. Long-term evaluation requires compressed and generalized causal structure, whereas short-term evaluation often relies on high-resolution local features.
\end{itemize}

\begin{figure}[h]
\centering
\begin{tikzpicture}[
  every node/.style={align=center,font=\small},
  source/.style={draw,rounded corners,minimum width=3.1cm,minimum height=0.75cm},
  basis/.style={draw,rounded corners,minimum width=4.4cm,minimum height=0.9cm},
  >=Latex
]

\node[source] (R) at (0,0) {Reference $R$\\$(\beta_R)$};
\node[source] (E) at (0,-1.25) {Exploration $E$\\$(H)$};
\node[source] (S) at (0,-2.50) {Stabilization $S$\\$(\tau)$};
\node[source] (D) at (0,-3.75) {Horizon $D$\\$(\gamma)$};

\node[basis] (ext) at (6.0,0) {Externalizability\\\scriptsize{(driven by $C_{\mathrm{obs}}, C_{\mathrm{coop}}$)}};
\node[basis] (of) at (6.0,-1.875) {Order / Freedom\\\scriptsize{(driven by $C_{\mathrm{comp}}, C_{\mathrm{coop}}$)}};
\node[basis] (ac) at (6.0,-3.75) {Abstract / Concrete\\\scriptsize{(driven by $C_{\mathrm{comp}}$)}};

\draw[->,thick,shorten >=2pt] (R.east) -- (ext.west);
\draw[->,thick,shorten >=2pt] (E.east) -- (of.west);
\draw[->,thick,shorten >=2pt] (S.east) -- (of.west);
\draw[->,thick,shorten >=2pt] (D.east) -- (ac.west);

\node[above=0.30cm of R,font=\itshape] {Inference profile $\theta$};
\node[above=0.30cm of ext,font=\itshape] {Three bases};

\end{tikzpicture}
\caption{Projection of the four-component inference profile $\theta = (R, E, S, D)$ onto three recurrent bases. Computational ($C_{\mathrm{comp}}$), observational ($C_{\mathrm{obs}}$), and coordination ($C_{\mathrm{coop}}$) constraints jointly induce the mapping: $R$ projects onto externalizability through the communication cost imposed by $C_{\mathrm{obs}}$ and $C_{\mathrm{coop}}$; $E$ and $S$ jointly project onto order/freedom through the plasticity--stability dilemma sharpened by reproducibility demands; $D$ projects onto abstract/concrete through rate--distortion compression under $C_{\mathrm{comp}}$.}
\label{fig:projection}
\end{figure}
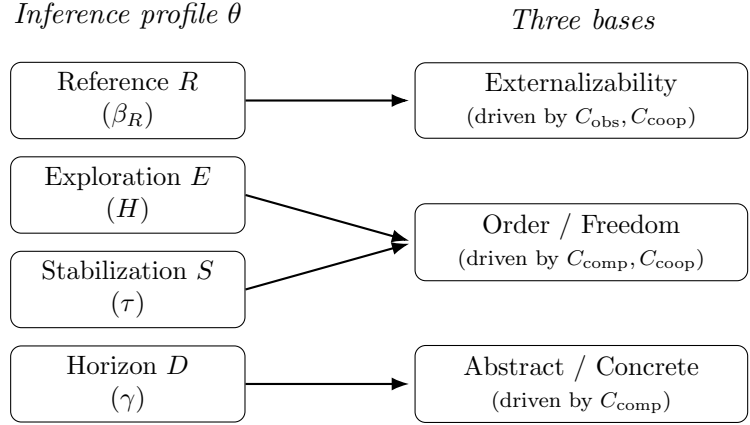

Figure~\ref{fig:projection} summarizes this projection.

Thus, although $\theta$ has four operational components, these components tend to reduce to three recurrent axes under finite resources. The reduction is not asserted as a strict theorem in this paper, but as a structural tendency supported by rate--distortion theory, communication cost, and the plasticity--stability dilemma. These three bases can be interpreted information-theoretically in a unified manner: abstract/concrete corresponds to rate--distortion trade-offs, externalizability corresponds to communication cost (description length), and order/freedom corresponds to the entropy of permitted diversity.

\section{Structural Correspondence in Deep Learning}

\subsection{Representation Hierarchy and Reference Selection}

In multilayer networks, lower layers often retain local and concrete features, whereas higher layers integrate more abstract representations~\cite{bengio2013}. Which layer or representation is used as the basis for readout is a design and operational degree of freedom. This provides a structural origin for the abstract/concrete axis.

In Transformers~\cite{vaswani2017}, residual streams can preserve concrete information across layers, while attention and multilayer transformations construct higher-order representations~\cite{elhage2021}. Even when the same input is processed, the decisive readout can depend on which representational level is operationally emphasized.

\subsection{Latent-State Estimation and Externalizability}

Under partial observability, hidden states cannot be uniquely reconstructed from observation $o$. This is not merely a coordination problem: it reflects a fundamental property of latent-variable models. Recent work on disentangled representation learning shows that, without inductive biases or supervision, the underlying generative factors are not identifiable from observations alone~\cite{locatello2019}, which can be read as a deep-learning instance of $W$-level non-identifiability. Sharing such states with other agents therefore requires interpretability, probing, logging, or other externalization procedures. In deep learning systems, layer activations, probe-readable representations, and audit logs are relatively externalizable signals, whereas high-dimensional latent states can be expressive but costly to communicate.

The parameter $\beta_R$ can therefore be interpreted as the degree to which inference relies on externalizable signals rather than high-description-cost latent representations.

\subsection{Regularization and Exploration}

Learning stability through weight decay, dropout, early stopping, or other regularization mechanisms trades off against adaptation and diversity. At inference time, sampling temperature and related parameters control the diversity of generated hypotheses. These mechanisms correspond to the order/freedom axis at the $\theta$ level: one may prioritize procedural reproducibility and stable outputs, or maintain broader alternative hypotheses and adaptive variation.

\section{Conditions for Identifiability}

\subsection{Resolving \texorpdfstring{$\theta$}{theta}-Level Non-Identifiability}

If divergence occurs under the same world model $W_\phi$, it can be reduced by explicitly synchronizing the dominant components of the inference profile, such as reference $R$ or horizon $D$. In an idealized case, alignment of operational settings yields
\begin{equation}
\mathrm{Infer}(W_\phi, o; \theta_A) \approx \mathrm{Infer}(W_\phi, o; \theta_B).
\label{eq:theta-align}
\end{equation}
This procedure identifies the disagreement not as a defect of personality but as a difference in inference operation.

\subsection{Reducing \texorpdfstring{$W$}{W}-Level Non-Identifiability}

If world models have diverged through long-term learning, merely aligning $\theta$ may not be sufficient. In that case, one must design discriminative observations $o^*$ or interventions $a$ that maximize predictive differences between the models:
\begin{equation}
|p_A(y \mid o^*) - p_B(y \mid o^*)| > \delta,
\label{eq:obs-discrim}
\end{equation}
for additional observation, or
\begin{equation}
|p_A(y \mid \mathrm{do}(a)) - p_B(y \mid \mathrm{do}(a))| > \delta,
\label{eq:int-discrim}
\end{equation}
for intervention~\cite{pearl2009}. Here $\mathrm{do}(a)$ denotes a causal intervention. The three bases provide useful coordinates for designing such discriminative observations.

\subsection{Implications of the Three Bases}

If the disagreement concerns externalizability, agents must pay the cost of making grounds mutually auditable. If it concerns order/freedom, exploration conditions such as temperature, test environments, or update thresholds must be aligned. If it concerns abstract/concrete orientation, evaluation horizons and levels of representation must be synchronized. These operations turn apparently non-identifiable disagreement into a model-adjustment problem involving additional observations, externalization, or intervention.

\section{Case Study: Two-Level Non-Identifiability in AI Regulation Debates}

\subsection{Shared Observations}

AI regulation debates provide a clear example of conclusion divergence under shared observations. Stakeholders often refer to shared materials: incident reports, misinformation cases, technical mitigation proposals, benchmark results, economic forecasts, and legal drafts. During the formation of the EU AI Act~\cite{eu_ai_act_2024}, for example, technical benchmarks, civil society reports, and industry impact assessments were all part of the public evidence environment. Yet the same materials supported divergent conclusions such as prohibition, risk-based regulation, phased regulation, or voluntary governance~\cite{veale2021,smuha2021}.

\subsection{\texorpdfstring{$\theta$}{theta}-Level Divergence}

The divergence can be represented as differences in $\theta = (R, E, S, D)$. Table~\ref{tab:ai-reg} summarizes a stylized contrast between precautionary and promotion-oriented positions. Precautionary actors often give weight to difficult-to-externalize concerns such as uncontrollability, misuse potential, and institutional fragility, retain worst-case scenarios as live alternatives, and emphasize long-term irreversibility. Promotion-oriented actors often give weight to externalizable benefits such as benchmarks and economic statistics, concentrate on mainline scenarios, and emphasize medium-term opportunity costs.

\begin{table}[h]
\centering
\caption{Stylized differences in inference profile $\theta$ in AI regulation debates. The contrast is schematic: actual stakeholder positions vary continuously along these axes rather than partitioning into two discrete types, and individual actors typically combine precautionary and promotion-oriented settings across different components.}
\label{tab:ai-reg}
\small
\begin{tabular}{@{}p{2.5cm}p{5.5cm}p{5.5cm}@{}}
\toprule
Component & Precautionary orientation & Promotion-oriented orientation \\
\midrule
Reference $R$ & Gives weight to hard-to-externalize concerns such as uncontrollability, misuse potential, and institutional fragility; smaller $\beta_R$. & Prioritizes externalizable grounds such as benchmarks, economic statistics, and observed benefits; larger $\beta_R$. \\
Exploration $E$ & Retains alternative hypotheses and worst-case scenarios; larger $H$. & Concentrates on mainline scenarios and local experimentation; smaller $H$. \\
Stabilization $S$ & Favors institutional fixation and high update thresholds; larger $\tau$. & Allows situational adaptation and flexible revision; smaller $\tau$. \\
Horizon $D$ & Emphasizes long-term irreversibility and loss of social trust; larger $\gamma$. & Emphasizes medium-term benefits and opportunity costs of delay; medium $\gamma$. \\
\bottomrule
\end{tabular}
\end{table}

\subsection{Signs of \texorpdfstring{$W$}{W}-Level Divergence}

Agents who have mainly learned from the history of technological innovation and benefit realization tend to assign large causal weight to the sequence
\begin{equation}
\text{early adoption} \to \text{trial and error} \to \text{improvement}.
\label{eq:promotive-causal}
\end{equation}
Agents who have mainly learned from histories of accidents, institutional failure, or distrust tend to assign large causal weight to the sequence
\begin{equation}
\text{absence of regulation} \to \text{accumulation of accidents} \to \text{loss of trust}.
\label{eq:cautious-causal}
\end{equation}
In this case, disagreement is not merely about which evidence is emphasized at the moment. The same new report may be interpreted as evidence of manageable iteration by one model and as evidence of systemic fragility by another, because the underlying causal models have diverged.

\subsection{Discriminative Observations and Interventions}

If the central issue is externalizability, progress requires operationalizing otherwise internalized concerns, such as by constructing measurable indicators of uncontrollability or institutional trust erosion. If the issue is order/freedom, limited-environment experiments or A/B tests that bound risk while preserving learning may serve as interventions~\cite{stilgoe2018}. The framework therefore shifts disagreement from defect attribution to the design of additional observations and interventions.

\section{Conclusion}

This paper formulated the phenomenon that conclusions diverge from the same observations as a two-level problem of non-identifiability. At the $\theta$ level, divergence arises from differences in inference operation: reference, exploration, stabilization, and horizon. At the $W$ level, repeated inference operations bias data exposure and updates, causing world models themselves to diverge.

We further argued that, under computational, observational, and coordination constraints, the four operational components of $\theta$ tend to be projected onto three recurrent bases: abstract/concrete, externalizability, and order/freedom. These bases are not arbitrary typologies; they can be interpreted through rate--distortion trade-offs, communication cost, and entropy or plasticity--stability trade-offs. By connecting the framework to deep representation learning, latent-state estimation, and regularization--exploration trade-offs, we positioned it as a computational account of disagreement rather than a purely rhetorical or psychological description.

Future work includes extracting inference profiles from empirical data such as dialogue and decision records, quantitative validation of the three-basis reduction, and formal design of discriminative observations for reducing world-model non-identifiability. Potential applications include mediation mechanisms in multi-agent systems, perspective-switching mechanisms in LLM-based dialogue systems, analysis of social consensus formation, and AI alignment under plural inference profiles.

\end{document}